\documentclass[11pt,a4paper]{article}
\usepackage[hyperref]{naaclhlt2019}

\usepackage{times}
\usepackage{latexsym}
\usepackage[linesnumbered,ruled]{algorithm2e}
\usepackage{url}
\usepackage{amsmath}
\usepackage{mathtools}
\usepackage{xcolor}
 \usepackage{booktabs}
\usepackage{mathrsfs}
\usepackage{multicol}

\usepackage{float}

\DeclareMathOperator{\RR}{\mathbb{R}}
\DeclareMathOperator*{\id}{\mathbf{I}}

\usepackage{amsfonts}

\aclfinalcopy % Uncomment this line for the final submission
\def\aclpaperid{1555} %  Enter the acl Paper ID here

\title{Continual Learning for Sentence Representations Using Conceptors}

\author{Tianlin Liu \\
  Department of Computer Science and \\ Electrical Engineering\\
Jacobs University Bremen \\
28759 Bremen, Germany\\
\texttt{t.liu@jacobs-university.de} \\
  \\\And
  Lyle Ungar \and Jo\~{a}o Sedoc\\
Department of Computer and \\ Information Science\\
University of Pennsylvania\\
Philadelphia, PA 19104\\
\texttt{\{ungar, joao\}@cis.upenn.edu}}

\date{}

\begin{document}
\maketitle
\begin{abstract}
Distributed representations of sentences have become ubiquitous in natural language processing tasks. In this paper, we consider a continual learning scenario for sentence representations: Given a sequence of corpora, we aim to optimize the sentence encoder with respect to the new corpus while maintaining its accuracy on the old corpora. To address this problem, we propose to initialize sentence encoders with the help of corpus-independent features, and then sequentially update sentence encoders using Boolean operations of {\it conceptor} matrices to learn corpus-dependent features. We evaluate our approach on semantic textual similarity tasks and show that our proposed sentence encoder can continually learn features from new corpora while retaining its competence on previously encountered corpora.
\end{abstract}

\section{Introduction}

Distributed representations of sentences are essential for a wide variety of natural language processing (NLP) tasks. Although recently proposed sentence encoders have achieved remarkable results (e.g., \citep{Yin2015, Arora2017, Cer2018, Pagliardini2018}), most, if not all, of them are trained on \emph{a priori} fixed corpora. However, in open-domain NLP systems such as conversational agents, oftentimes we are facing a dynamic environment, where training data are accumulated sequentially over time and the distributions of training data vary with respect to external input \citep{Lee2017, Mathur2018}. To effectively use sentence encoders in such systems, we propose to consider the following \emph{continual sentence representation learning task}: Given a sequence of corpora, we aim to train sentence encoders such that they can continually learn features from new corpora while retaining strong performance on previously encountered corpora. 

Toward addressing the continual sentence representation learning task, we propose a simple sentence encoder that is based on the summation and linear transform of a sequence of word vectors aided by matrix conceptors. Conceptors have their origin in reservoir computing \cite{Jaeger2014} and recently have been used to perform continual learning in deep neural networks \cite{He2018}. Here we employ Boolean operations of conceptor matrices to update sentence encoders over time to meet the following desiderata:

\begin{enumerate} 
\item \emph{Zero-shot learning}. The initialized sentence encoder (no training corpus used) can effectively produce sentence embeddings.
\item \emph{Resistant to catastrophic forgetting}. When the sentence encoder is adapted on a new training corpus, it retains strong performances on old ones.
\end{enumerate}

The rest of the paper is organized as follows. We first briefly review a family of linear sentence encoders. Then we explain how to build upon such sentence encoders for continual sentence representation learning tasks, which lead to our proposed algorithm. Finally, we demonstrate the effectiveness of the proposed method using semantic textual similarity tasks.\footnote{Our codes are available on GitHub \url{https://github.com/liutianlin0121/contSentEmbed}}

\paragraph{Notation}
We assume each word $w$ from a vocabulary set $V$ has a real-valued word vector $v_w \in \mathbb{R}^n$. Let $p(w)$ be the monogram probability of a word $w$. A corpus $D$ is a collection of sentences, where each sentence $s \in D$ is a multiset of words (word order is ignored here). For a collection of vectors $Y = \{y_i\}_{i \in I}$, where $y_i \in \RR^l$ for $i$ in an index set $I$ with cardinality $|I|$, we let $[y_i]_{i \in I} \in \RR^{l \times |I|}$ be a matrix whose columns are vectors $y_1, \cdots, y_{|I|}$. An identity matrix is denoted by $\mathbf{I}$.

\section{Linear sentence encoders}
We briefly overview ``linear sentence encoders'' that are based on linear algebraic operations over a sequence of word vectors. Among different linear sentence encoders, the smoothed inverse frequency (SIF) approach \citep{Arora2017} is a prominent example -- it outperforms many neural-network based sentence encoders on a battery of NLP tasks \citep{Arora2017}.

Derived from a generative model for sentences, the SIF encoder (presented in Algorithm \ref{alg:sif}) transforms a sequence of word vectors into a sentence vector with three steps. First, for each sentence in the training corpus, SIF computes a weighted average of word vectors (line 1-3 of Algorithm \ref{alg:sif}); next, it estimates a ``common discourse direction'' of the training corpus (line 4 of Algorithm \ref{alg:sif}); thirdly, for each sentence in the testing corpus, it calculates the weighted average of the word vectors and projects the averaged result away from the learned common discourse direction (line 5-8 of Algorithm \ref{alg:sif}). Note that this 3-step paradigm is slightly more general than the original one presented in \citep{Arora2017}, where the training and the testing corpus is assumed to be the same.

\begin{algorithm}%[htbp]
\SetKwInOut{Input}{Input}
\SetKwInOut{Output}{Output}
\Input{A training corpus $D$; a testing corpus $G$; parameter $a$,  monogram probabilities $\{p(w)\}_{w \in V}$ of words}
\For{sentence $s \in D$}{$q_s \leftarrow \frac{1}{ |s|} \sum\limits_{w \in s} \frac{a}{p(w) + a} v_w$}
Let $u$ be the first singular vector of $[q_s]_{s \in D}$. \\
\For{sentence $s \in G$}{ 
$q_s \leftarrow \frac{1}{ |s|} \sum\limits_{w \in s} \frac{a}{p(w) + a} v_w$ \\
$f^{\text{SIF}}_s  \leftarrow  q_s - u u^{\top} q_s$.}
\Output{$\{f^{\text{SIF}}_s\}_{s \in G}$} 
\caption{SIF sentence encoder.}
\label{alg:sif}
\end{algorithm}

%In this work, when we talk about ``training a SIF encoder,'' we refer to the estimation of common discourse $u$ (line 1-4) using a training corpus $D$. With an estimated $u$, we can apply SIF to embed any sentence (not necessarily in $D$), by re-using line 2 and 6 of Algorithm \ref{alg:sif}. 

%Empirically, \citet{Arora2017} showed that SIF outperforms many sophisticated, neural-network-based sentence encoder when tested on a battery of textual similarity tasks and downstream supervised learning tasks.

Building upon SIF, recent studies have proposed further improved sentence encoders \citep{Khodak2018, Pagliardini2018, Yang2018}. These algorithms roughly share the core procedures of SIF, albeit using more refined methods (e.g., softly remove more than one common discourse direction).

%Building upon SIF sentence encoder, recent studies have proposed further improved sentence vector algorithms. For example, \citet{Khodak2018} proposed to use a learned linear transformation to remove common discourse information; \citet{Pagliardini2018} proposed to remove common discourse information softly; \citet{Yang2018} leverage Gram-Schmidt Process before common discourse removal. These algorithms, although have subtle differences, roughly share two steps employed by SIF: (1) perform a weighted average of word vectors, and (2) use a linear transformation to remove common discourse information shared in a corpus.

\section{Continual learning for linear sentence encoders}

In this section, we consider how to design a linear sentence encoder for continual sentence representation learning.  We observe that common discourse directions used by SIF-like encoders are estimated from the training corpus. However, incrementally estimating common discourse directions in continual sentence representation learning tasks might not be optimal. For example, consider that we are sequentially given training corpora of \texttt{tweets} and \texttt{news article}. When the first \texttt{tweets} corpus is presented, we can train a SIF sentence encoder using \texttt{tweets}. When the second \texttt{news article} corpus is given, however, we will face a problem on how to exploit the newly given corpus for improving the \emph{trained} sentence encoder. A straightforward solution is to first combine the \texttt{tweets} and \texttt{news article} corpora and then train a new encoder from scratch using the combined corpus. However, this paradigm is not efficient or effective. It is not efficient in the sense that we will need to re-train the encoder from scratch every time a new corpus is added. Furthermore, it is not effective in the sense that the common direction estimated from scratch reflects a compromise between tweets and news articles, which might not be optimal for either of the stand-alone corpus. Indeed, it is possible that larger corpora will swamp smaller ones.%; it is therefore common in transfer learning and continual learning to do ad hoc weighting of the different corpora.

To make the common discourse learned from one corpus more generalizable to another, we propose to use the conceptor matrix \citep{Jaeger2017} to characterize and update the common discourse features in a sequence of training corpora. 

\subsection{Matrix conceptors}

In this section, we briefly introduce matrix conceptors, drawing heavily on  \citep{Jaeger2017, He2018, Liu2019}. Consider a set of vectors  $\{x_1, \cdots, x_n\}$,  $x_i \in \mathbb{R}^N$ for all $i \in \{1, \cdots, n\}$. A conceptor matrix is a regularized identity map that minimizes 
\begin{equation}
\label{matconceptor}
\frac{1}{n} \sum_{i=1}^{n}  \|x_i - C x_i\|_2^2+\alpha^{-2}\|C\|_{\text{F}}^2.
\end{equation}
where $\| \cdot \|_{\text{F}}$ is the Frobenius norm and $\alpha^{-2}$ is a scalar parameter called \emph{aperture}. It can be shown that $C$ has a closed form solution: 
\begin{equation}
\label{conceptorsolution}
C = \frac{1}{n} X X^{\top} (\frac{1}{n} X X^{\top}+\alpha^{-2} I)^{-1},
\end{equation}
where $X = [x_i]_{i \in \{1, \cdots, n\}}$ is a data collection matrix whose columns are vectors from $\{x_1, \cdots, x_n\}$. In intuitive terms, $C$ is a soft projection matrix on the linear subspace where the typical components of $x_i$ samples lie. For convenience in notation, we may write $C (X, \alpha)$ to stress the dependence on $X$ and $\alpha$. 

Conceptors are subject to most laws of Boolean logic such as NOT $\neg$, AND $\wedge$ and OR $\vee$. For two conceptors $C$ and $B$, we define the following operations:
\begin{align}
\label{bool}
\neg C:=& \id-C,\\
C\wedge B:=&  (C^{-1} + B^{-1} - \id)^{-1} \\
C \vee B:=&\neg(\neg C \wedge \neg B) 
\end{align}

Among these Boolean operations, the OR operation $\vee$ is particularly relevant for our continual sentence representation learning task. It can be shown that $C \vee B$ is the conceptor computed from the union of the two sets of sample points from which $C$ and $B$ are computed. Note that, however, to calculate $C \vee B$, we only need to know two matrices $C$ and $B$ and do not have to access to the two sets of sample points from which $C$ and $B$ are computed.
 
\subsection{Using conceptors to continually learn sentence representations} \label{sec:algorithm}

We now show how to sequentially characterize and update the common discourse of corpora using the Boolean operation of conceptors. Suppose that we are sequentially given $M$ training corpora $D^1, \cdots, D^M$, presented one after another. Without using any training corpus, we first initialize a conceptor which characterizes the corpus-independent common discourse features. More concretely, we compute $C^0 \coloneqq C ( [v_w]_{w \in Z}, \alpha)$, where $[v_w]_{w \in Z}$ is a matrix of column-wisely stacked word vectors of words from a stop word list $Z$ and $\alpha$ is a hyper-parameter. After initialization, for each new training corpus $D^i$ ($i = 1, \cdots, M$) coming in, we compute a new conceptor $C^{\text{temp}} \coloneqq C ( [q_s]_{s \in D^i}, \alpha)$ to characterize the common discourse features of corpus $D^i$, where those $q_s$ are defined in the SIF Algorithm \ref{alg:sif}. We can then use Boolean operations of conceptors to compute $C^i \coloneqq C^{\text{temp}} \vee C^{i-1}$, which characterizes common discourse features from the new corpus as well as the old corpora. After all $M$ corpora are presented, we follow the SIF paradigm and use $C^M$ to remove common discourse features from (potentially unseen) sentences. The above outlined conceptor-aided (CA) continual sentence representation learning method is presented in Algorithm \ref{algo:conceptorAlg}.

\begin{algorithm}[htbp]
\SetKwInOut{Input}{Input}
\SetKwInOut{Output}{Output}
\Input{A sequence of $M$ training corpora $\mathcal{D} = \{ D^1, \cdots, D^M\}$; a testing corpus $G$; hyper-parameters $a$ and $\alpha$; word probabilities $\{p(w)\}_{w \in V}$; stop word list $Z$.}
$C^0 \leftarrow C ( [v_w]_{w \in Z}, \alpha)$ . \\
\For{corpus index $i = 1, \cdots, M$}{
\For{sentence $s \in D^i$}{$q_s \leftarrow \frac{1}{ |s|} \sum\limits_{w \in s} \frac{a}{p(w) + a} v_w$}
$C^{\text{temp}} \leftarrow C ( [q_s]_{s \in D^i}, \alpha)$ \\
$C^i \leftarrow C^{\text{temp}} \vee C^{i-1}$}
\For{$s \in G$}{
$q_s \leftarrow \frac{1}{ |s|} \sum\limits_{w \in s} \frac{a}{p(w) + a} v_w$ \\
$f^{\text{CA}}_s  \leftarrow  q_s - C^M q_s$}
\Output{$\{f^{\text{CA}}_s\}_{s \in G}$}
\caption{CA sentence encoder.}
\label{algo:conceptorAlg}
\end{algorithm}

A simple modification of Algorithm \ref{algo:conceptorAlg} yields a ``zero-shot'' sentence encoder that requires only pre-trained word embeddings and no training corpus: we can simply skip those corpus-dependent steps (line 2-8) and use $C^0$ in place of $C^M$ in line 11 in Algorithm \ref{algo:conceptorAlg} to embed sentences. This method will be referred to as ``zero-shot CA.''

\begin{figure*}[ht]
%\centering
\includegraphics[width = \textwidth]{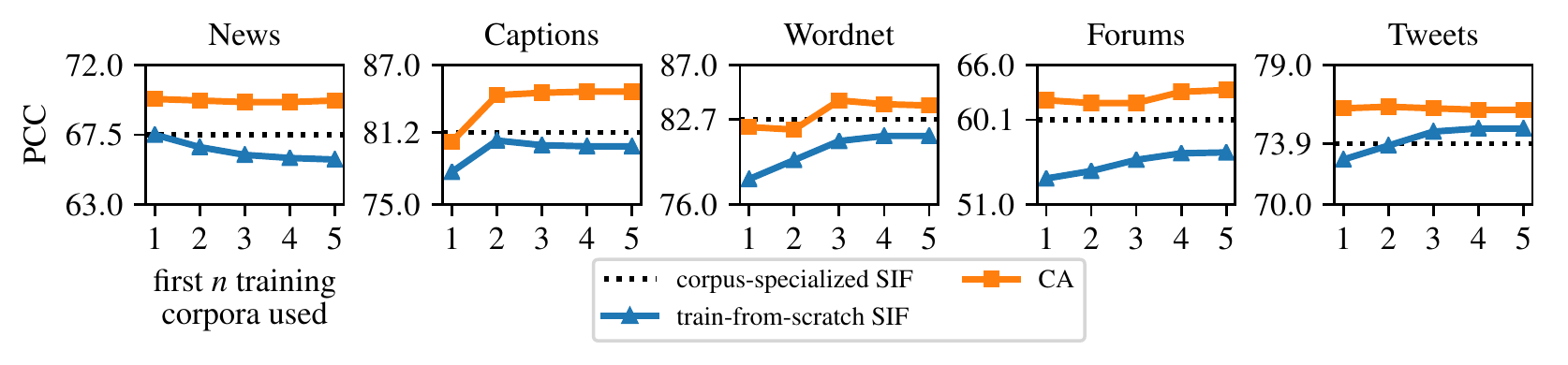}
\caption{PCC results of STS datasets. Each panel shows the PCC results of a testing corpus (specified as a subtitle) as a function of increasing numbers of training corpora used. The setup of this experiment mimics \cite[section 5.1]{Zenke2017}.}
\label{fig:result}
\end{figure*}

\begin{table*}[ht] 
\centering
\begin{tabular}{r  c c c c c}
\toprule
                  &  News & Captions & WordNet & Forums & Tweets  \\ \midrule
%corpus specialized SIF  &     47.2 & 60.1 & 82.7 & 81.2  &  67.5\\ 
av. train-from-scratch SIF  &  \underline{66.5} & 79.7 & 80.3 & 55.5 & 74.2 \\ 
zero-shot CA  &    65.6 & \underline{79.8} & \underline{82.5} & \underline{61.5} & \underline{75.2} \\ 
av. CA               & \bf 69.7 & \bf 83.8 & \bf 83.2 & \bf 62.5 & \bf 76.2\\  \bottomrule
\end{tabular}
\caption{Time-course averaged PCC of train-from-scratch SIF and  conceptor-aided (CA) methods, together with the result of zero-shot CA. Best results are in boldface and the second best results are underscored.}
\label{tb:zeroShot}
\end{table*}

\section{Experiment}

We evaluated our approach for continual sentence representation learning using semantic textual similarity (STS) datasets \citep{Agirre2012,Agirre2013,Agirre2014,Agirre2015, Agirre2016}. The evaluation criterion for such datasets is the Pearson correlation coefficient (PCC) between the predicted sentence similarities and the ground-truth sentence similarities. We split these datasets into five corpora by their genre:  news, captions, wordnet, forums, tweets (for details see appendix). Throughout this section, we use publicly available 300-dimensional GloVe vectors (trained on the 840 billion token Common Crawl) \citep{Pennington2014}. Additional experiments with Word2Vec \citep{Mikolov2013}, Fasttext \citep{Bojanowski2017}, Paragram-SL-999 \citep{Wieting2015} are in the appendix. 

We use a standard continual learning experiment setup (cf. \citep[section 5.1]{Zenke2017})  as follows. We sequentially present the five training datasets in the order\footnote{The order can be arbitrary. Here we ordered the corpora from the one with the largest size (news) to the smallest size (tweets). The results from reversely ordered corpora are reported in the appendix.} of news, captions, wordnet, forums, and tweets, to train sentence encoders. Whenever a new training corpus is presented, we train a SIF encoder from scratch\footnote{We use $a = 0.001$ as in \citep{Arora2017}. The word frequencies are available at the GitHub repository of SIF.} (by combining all available training corpora which have been already presented) and then test it on each corpus. At the same time, we incrementally adapt a CA encoder\footnote{We used hyper-parameter $\alpha = 1$. Other parameters are set to be the same as SIF.} using the newly presented corpus and test it on each corpus. The lines of each panel of Figure \ref{fig:result} show the test results of SIF and CA on each testing corpus (specified as the panel subtitle) as a function of the number of training corpora used (the first $n$ corpora of news, captions, wordnet, forums, and tweets for this experiment). To give a concrete example, consider the blue line in the first panel of Figure \ref{fig:result}. This line shows the test PCC scores ($y$-axis) of SIF encoder on the news corpus when the number of training corpora increases ($x$-axis). Specifically, the left-most blue dot indicates the test result of SIF encoder on news corpus when trained on news corpus itself (that is, the first training corpus is used); the second point indicates the test results of SIF encoder on news corpus when trained on news and captions corpora (i.e., the first \emph{two} training corpora are used); the third point indicates the test results of SIF encoder on news corpus when trained on news, captions, and wordnet corpora (that is, the first \emph{three} training corpora are used), so on and so forth. The dash-lines in panels show the results of a corpus-specialized SIF, which is trained and tested on the same corpus, i.e., as done in \citep[section 4.1]{Arora2017}. We see that the PCC results of CA are better and more ``forgetting-resistant'' than train-from-scratch SIF throughout the time course where more training data are incorporated. Consider, for example, the test result of news corpus (first panel) again. As more and more training corpora are used, the performance of train-from-scratch SIF drops with a noticeable slope; by contrast, the performance CA drops only slightly. 

As remarked in the section \ref{sec:algorithm}, with a simple modification of CA, we can perform zero-shot sentence representation learning without using any training corpus. The zero-shot learning results are presented in Table \ref{tb:zeroShot}, together with the time-course averaged results of CA and train-from-scratch SIF (i.e., the averaged values of those CA or SIF scores in each panel of Figure \ref{fig:result}). We see that the averaged results of our CA method performs the best among these three methods. Somewhat surprisingly, the results yielded by zero-shot CA are better than the averaged results of train-from-scratch SIF in most of the cases.

We defer additional experiments to the appendix, where we compared CA against more baseline methods and use different word vectors other than GloVe to carry out the experiments. 

\section{Conclusions and future work}

In this paper, we formulated a continual sentence representation learning task: Given a consecutive sequence of corpora presented in a time-course manner, how can we extract useful sentence-level features from new corpora while retaining those from previously seen corpora? We identified that the existing linear sentence encoders usually fall short at solving this task as they leverage on ``common discourse'' statistics estimated based on a priori fixed corpora. We proposed two sentence encoders (CA encoder and zero-shot CA encoder) and demonstrate their the effectiveness at the continual sentence representation learning task using STS datasets.

As the first paper considering continual sentence representation learning task, this work has been limited in a few ways -- it remains for future work to address these limitations. First, it is worthwhile to incorporate more benchmarks such as GLUE \citep{Wang2018GLUE} and SentEval \citep{Conneau2018} into the continual sentence representation task. Second, this work only considers the case of linear sentence encoder, but future research can attempt to devise (potentially more powerful) non-linear sentence encoders to address the same task. Thirdly, the proposed CA encoder operates at a corpus level, which might be a limitation if boundaries of training corpora are ill-defined. As a future direction, we expect to lift this assumption, for example, by updating the common direction statistics at a sentence level using Autoconceptors \citep[section 3.14]{Jaeger2014}. Finally, the continual learning based sentence encoders should be applied to downstream applications in areas such as open domain NLP systems.

%Given the results of our sentence encoders for semantic textual similarity, we will apply our method to text simplification and dialog generation.

\section*{Acknowledgement}
The authors thank anonymous reviewers for their helpful feedback. This work was partially supported by Jo\~{a}o Sedoc's Microsoft Research Dissertation Grant.

\bibliography{nlp_progress.bib}

\clearpage
\setcounter{page}{1}
\onecolumn

\part*{Supplementary Information}

\section*{The split STS datasets}

In the main body of the paper, we have reported that we have used the STS datasets split by genre. A detailed list such STS tasks can be found in Table \ref{tb:sts:tasks} and can be downloaded from \url{http://ixa2.si.ehu.es/stswiki/index.php/STSbenchmark} and \url{http://ixa2.si.ehu.es/stswiki/index.php/STSbenchmark#Companion}.

\begin{table}[H]
\centering
\scalebox{0.75}{
\begin{tabular}{ r c c c c}
\toprule
News & Captions & Forum & Tweets & WN  \\
\midrule
MSRpar 2012 & MSRvid 2012 & deft-forum 2014 & tweet-news 2014 &  OnWN 2012-2014 \\
headlines 2013-2016 & images 2014-2015 & answers-forums 2015 &  &  \\
deft-news 2014 & track5.en-en 2017 & answer-answer  2016 &  &  \\ \midrule
4299 sentence pairs & 3250 sentence pairs & 1079 sentence pairs&  750 sentence pairs& 2061 sentence pairs\\
\bottomrule
\end{tabular}}
\caption{STS datasets breakdown according to genres.}
\label{tb:sts:tasks}
\end{table}

\section*{CA compared with incremental-deletion SIF}

\begin{figure}[H]
%\centering
\includegraphics[width = \textwidth]{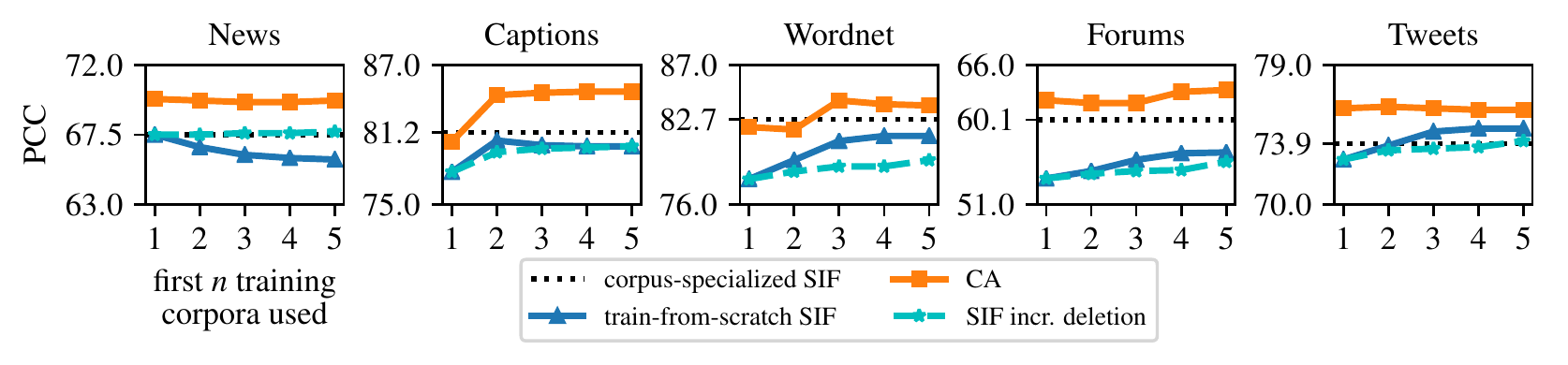}
\caption{Pearson correlation coefficients (PCC) of the split STS datasets as a function of the number of training corpora. For explanation see text.}
\label{fig:inc_sif}
\end{figure}

We compare the CA approach with the following variant of SIF. In the learning phase, for each corpus coming in, we learn and store a common direction (estimated based on the new corpus). In the testing phase, for a sentence in the testing corpora, we project it away from all common directions we have stored so far. We call this approach SIF with incremental deletions. The testing result is reported in Figure \ref{fig:inc_sif}.

\section*{CA without stop word initialization}

\begin{figure}[H]
%\centering
\includegraphics[width = \textwidth]{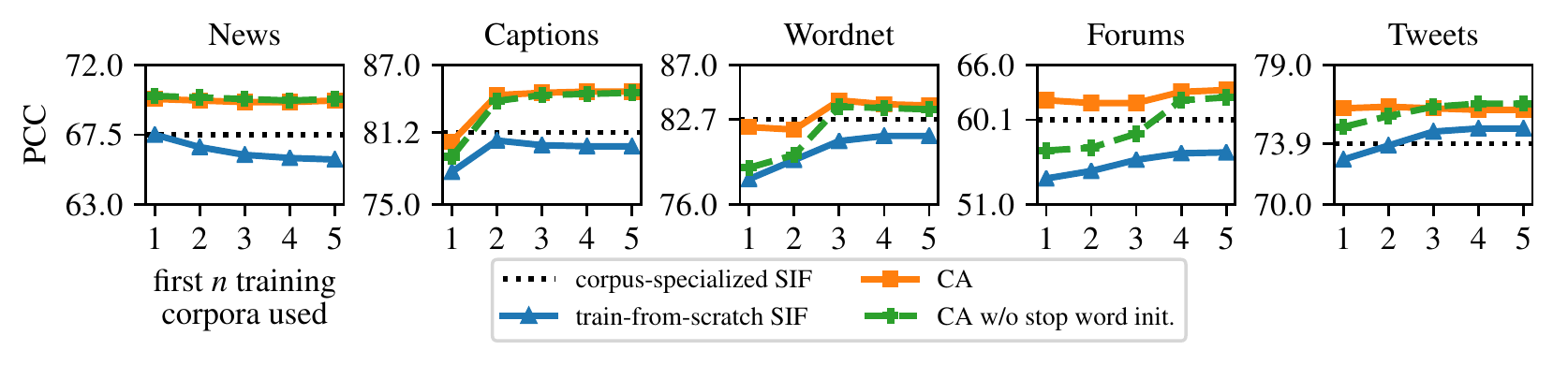}
\caption{Pearson correlation coefficients (PCC) of the split STS datasets as a function of the number of training corpora. For explanation see text.}
\label{fig:result_wo_stopwords}
\end{figure}

We have also tested the performance of CA without the initializing our concepor $C^0$ by stop words. That is, we set $C^0$ as a zero matrix in our CA algorithm. The results are reported in Figure \ref{fig:result_wo_stopwords}

We see that, the CA initialized by stop words are more beneficial than without such initializations, especially for those testing corpora that are unseen in training data.

\section*{CA with the reverse-ordered sequence of training corpora}

\begin{figure}[H]
%\centering
\includegraphics[width = \textwidth]{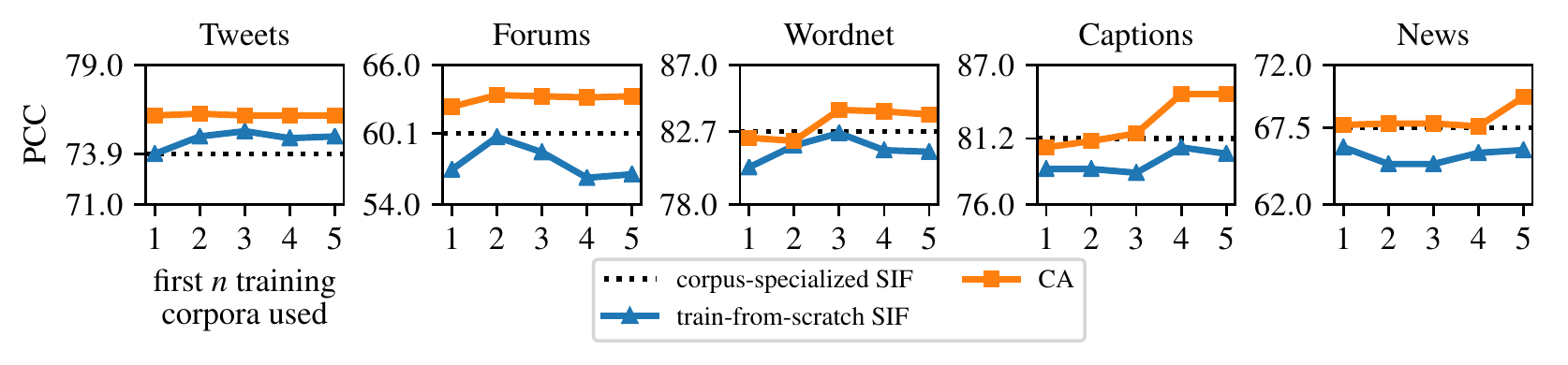}
\caption{Pearson correlation coefficients (PCC) of the split STS datasets as a function of the number of training corpora. For explanation see text.}
\label{fig:result2}
\end{figure}

In the main body of the paper, we sequentially presented new training corpus for sentence encoders, from the corpora of largest size (news) to that of the smallest size (tweets). We have remarked that this choice of ordering is essentially arbitrary. We now report the results for the reverse order (i.e., from corpora of smallest size to that of largest size) in Figure \ref{fig:result2}. We see that CA approach still outperforms train-from-scratch SIF throughout the time course.

\section*{Experiment using other word embedding brands}

We repeat the experiments with Word2Vec \citep{Mikolov2013}\footnote{\url{https://code.google.com/archive/p/word2vec/}} (pre-trained on Google News; 3 million tokens), Fasttext \citep{Bojanowski2017} \footnote{\url{https://fasttext.cc/docs/en/english-vectors.html}} (pre-trained on Common Crawl; 2 million of tokens), and Paragram SL-999\footnote{\url{https://cogcomp.org/page/resource_view/106}} (fine-tuned based on GloVe). The pipeline of the experiments echo that of the main body of the paper.

\subsection*{Using Word2vec}

\begin{figure}[H]
%\centering
\includegraphics[width = \textwidth]{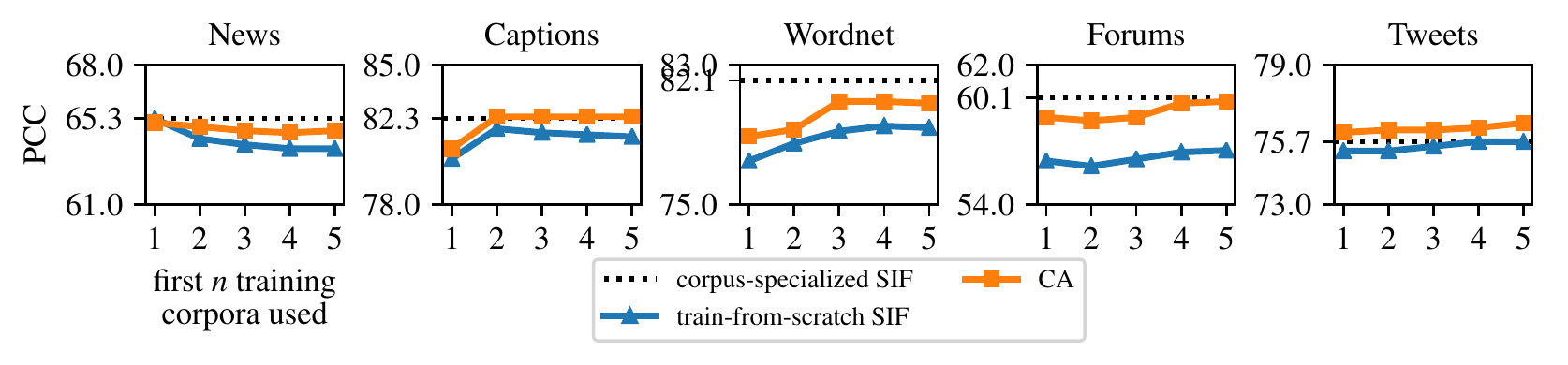}
\caption{Pearson correlation coefficients (PCC) of the split STS datasets as a function of the number of training corpora. Word2Vec is used.}
\label{fig:result_w2v}
\end{figure}

\subsection*{Using Fasttext}

\begin{figure}[H]
%\centering
\includegraphics[width = \textwidth]{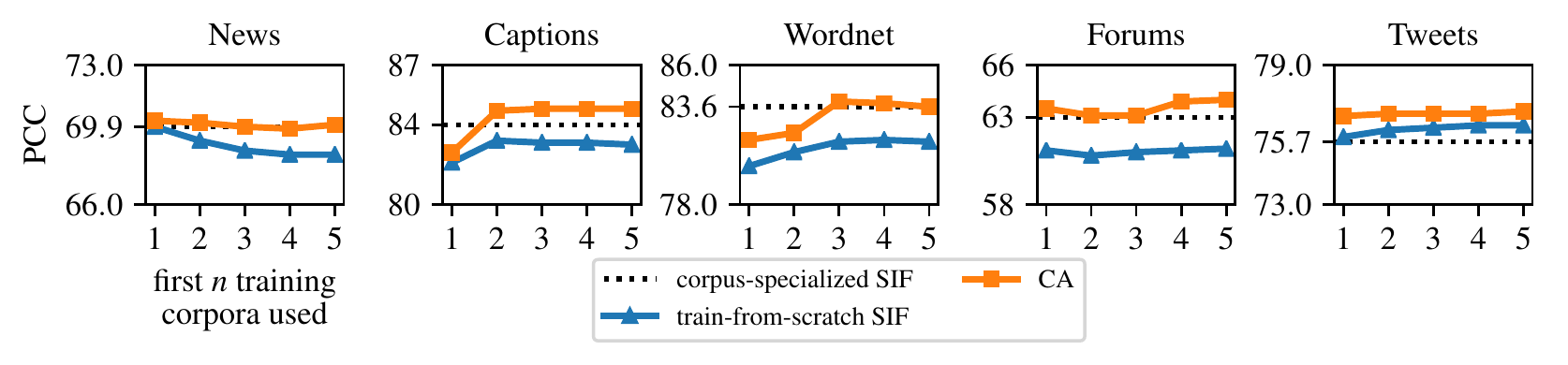}
\caption{Pearson correlation coefficients (PCC) of the split STS datasets as a function of the number of training corpora. Fasttext is used.}
\label{fig:result_fasttext}
\end{figure}

\subsection*{Using Paragram-SL-999}

\begin{figure}[H]
%\centering
\includegraphics[width = \textwidth]{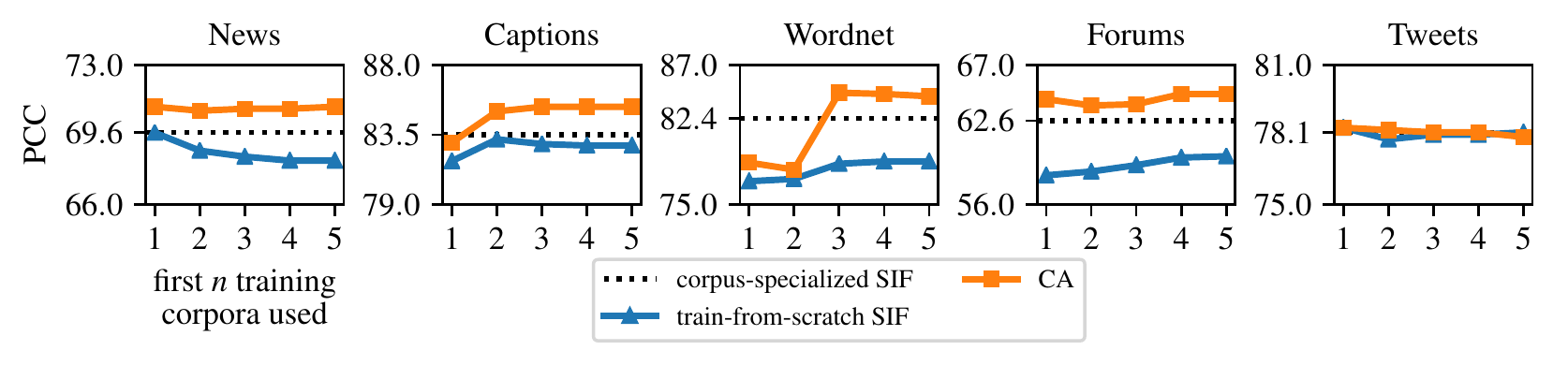}
\caption{Pearson correlation coefficients (PCC) of the split STS datasets as a function of the number of training corpora. Paragram-SL-999 is used.}
\label{fig:result_psl}
\end{figure}

\bibliographystyle{acl_natbib}

\end{document}